\documentclass{article}
\usepackage{ijcai25}

\usepackage{times}
\usepackage{tabularx}
\usepackage{xcolor}
\usepackage{soul}
\usepackage{url}
\usepackage[hidelinks]{hyperref}
\usepackage[utf8]{inputenc}
\usepackage[small]{caption}
\usepackage{graphicx}
\usepackage{amsmath}
\usepackage{enumerate}
\usepackage{amsthm}
\usepackage{thmtools}
\usepackage{amssymb}
\usepackage{multirow}
\usepackage{booktabs}
\usepackage{algorithm}
\usepackage{algorithmic}
\usepackage{subfigure}
\usepackage[switch]{lineno}
\declaretheoremstyle[
    headfont=\itshape\bfseries,%
    notefont=\bfseries, notebraces={}{},%
    bodyfont=\upshape,%
    postheadspace=1em
]{definitionstyle}

\declaretheorem[style=definitionstyle, name=Definition]{definition}

\title{KnowRA: Knowledge Retrieval Augmented Method for Document-level Relation Extraction with Comprehensive Reasoning Abilities}

\author{
Chengcheng Mai$^{1,2,3}$
\and
Yuxiang Wang$^{2,3}$\and
Ziyu Gong$^{2,3}$\and
Hanxiang Wang$^{2,3}$\And
Yihua Huang$^{2,3}$\\
\affiliations
$^1$ Nanjing Normal University, School of Computer Science and Electronic Informatics\\
$^2$ State Key Laboratory for Novel Software Technology at Nanjing University\\
$^3$ Nanjing University, School of Computer Science\\
\emails
\ maicc@njnu.edu.cn,
yuxiangwang@smail.nju.edu.cn, ziyugong@smail.nju.edu.cn, wanghanxiang@smail.nju.edu.cn, yhuang@nju.edu.cn}

\begin{document}

\maketitle

\begin{abstract}
    Document-level relation extraction (Doc-RE) aims to extract relations between entities across multiple sentences. Therefore, Doc-RE requires more comprehensive reasoning abilities like humans, involving complex cross-sentence interactions between entities, contexts, and external general knowledge, compared to the sentence-level RE. However, most existing Doc-RE methods focus on optimizing single reasoning ability, but lack the ability to utilize external knowledge for comprehensive reasoning on long documents. To solve these problems, a knowledge retrieval augmented method, named KnowRA, was proposed with comprehensive reasoning to autonomously determine whether to accept external knowledge to assist DocRE. Firstly, we constructed a document graph for semantic encoding and integrated the co-reference resolution model to augment the co-reference reasoning ability. Then, we expanded the document graph into a document knowledge graph by retrieving the external knowledge base for common-sense reasoning and a novel knowledge filtration method was presented to filter out irrelevant knowledge. Finally, we proposed the axis attention mechanism to build direct and indirect associations with intermediary entities for achieving cross-sentence logical reasoning. Extensive experiments conducted on two datasets verified the effectiveness of our method compared to the state-of-the-art baselines. Our code is available at https://anonymous.4open.science/r/KnowRA. This work has been accepted by IJCAI 2025.
\end{abstract}

\section{Introduction}

The document-level relation extraction (Doc-RE) task aims to extract pre-defined relation triples from documents containing multiple sentences. Compared with the existing sentence-level RE task, Doc-RE is not only more complicated but also more fundamental for real-world applications, like retrieval augmented generation (RAG) method for large language model (LLM) \cite{ref1,ref2,ref4}, automatic question and answering \cite{ref5,ref6}, and event extraction \cite{ref7,ref8,ref9}.

\begin{figure}[!htb]
    \centering
    \includegraphics[width=\linewidth]{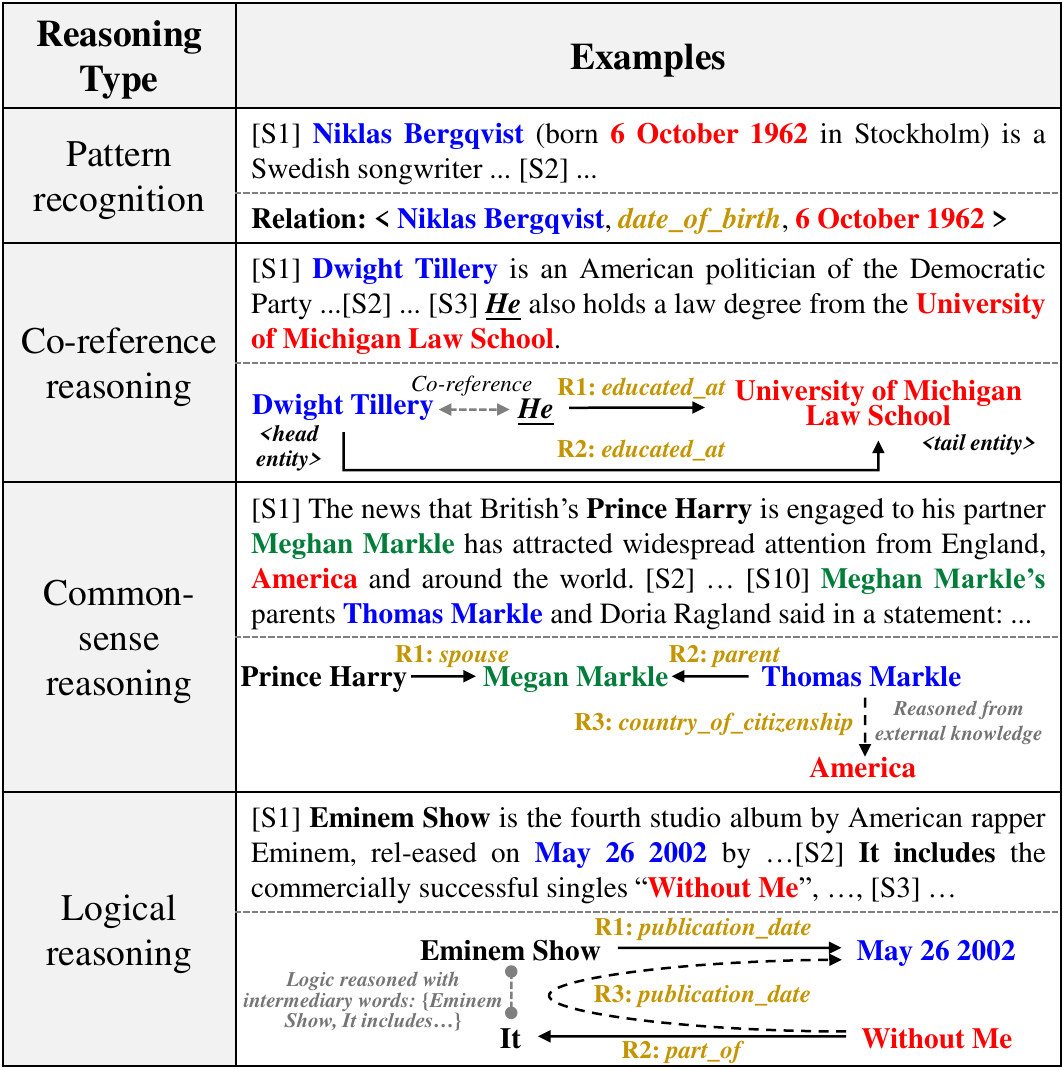}
    \caption{Different reasoning abilities for Doc-RE. Relation R3 in common-sense reasoning can hardly be extracted from the original document but can be retrieved from external knowledge. Relation R3 in logical reasoning can be reasoned by intermediary words.}
    \label{fig1}
\end{figure}

Obviously, Doc-RE models need to have comprehensive reasoning abilities for Doc-RE. Previous studies divided these abilities into four categories \cite{ref10}: pattern recognition, co-reference reasoning, common-sense reasoning, and logical reasoning, as shown in Figure \ref{fig1}. 

However, the existing Doc-RE models were mostly optimized for partial reasoning abilities and lack of comprehensive reasoning abilities. For pattern recognition, most recent methods constructed graph structures \cite{ref11,ref12,ref13,ref14} to establish long-distance associations between entities. For logical reasoning, several models used bridge/intermediary entities to establish indirect relations between two entities that are not directly connected \cite{ref15,ref16,ref17}. Considering that entity mentions often appear in the form of co-reference pronouns in the document, methods based on co-reference reasoning have been proposed \cite{ref18} and focus on identifying co-reference pronouns of entities, which were used as bridge entities to establish indirect relations between entities.

Also, the Doc-RE model requires the ability of common-sense reasoning because some relations cannot be inferred from the document itself, and external knowledge is needed to assist in Doc-RE. However, few existing Doc-RE models combined external knowledge for common-sense reasoning \cite{ref19}.

To summarize, the main challenges for Doc-RE lie in: 1) How to integrate the comprehensive reasoning ability required by Doc-RE. 2) How to represent and integrate external knowledge with the internal semantics of the document. 3) How to autonomously determine whether to accept external knowledge, considering that external knowledge may be lagging, one-sided, or even wrong.

\begin{figure*}
    \centering
    \includegraphics[width=0.9\textwidth]{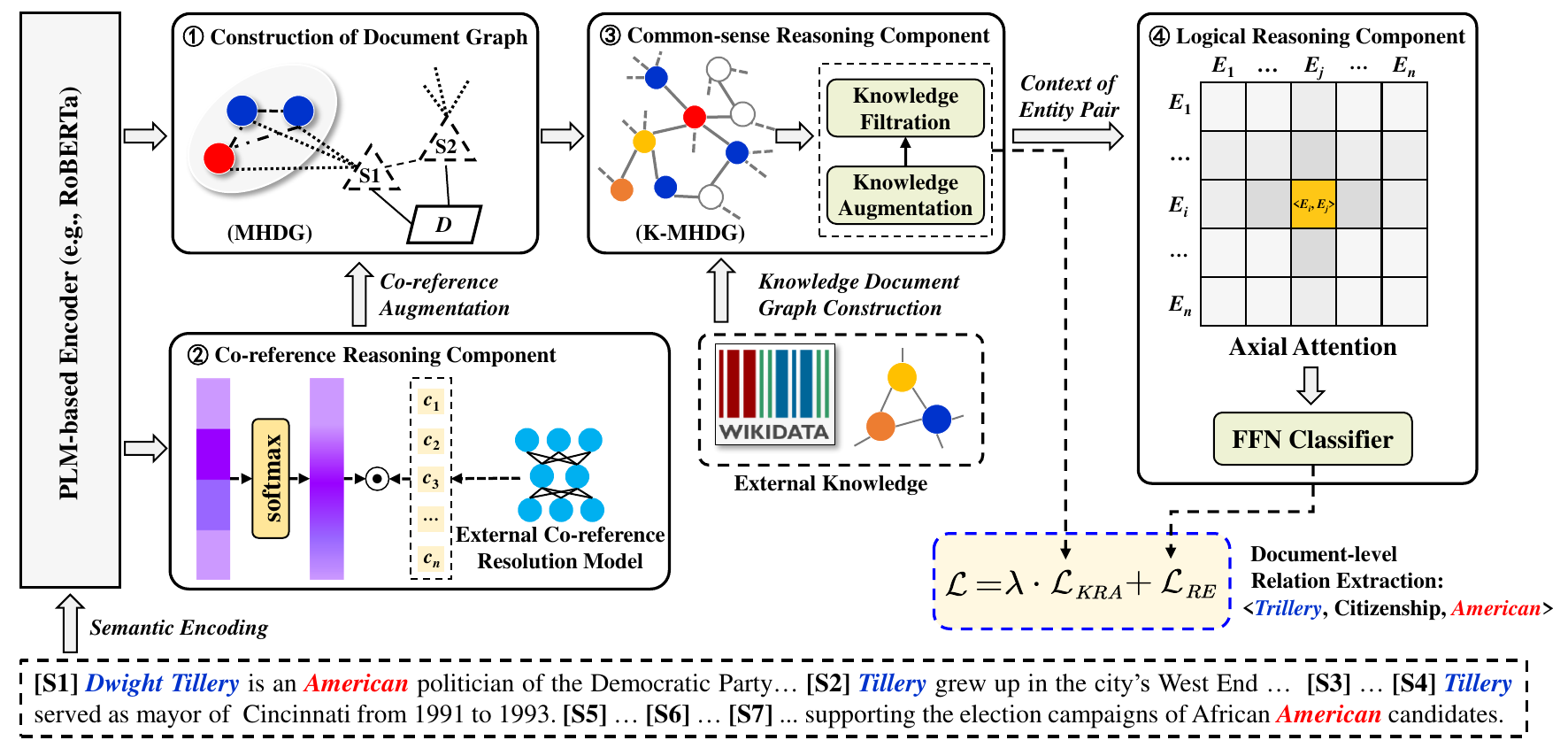}
    \caption{Overview of the proposed comprehensive reasoning method with knowledge augmentation for Doc-RE.}
    \label{fig2}
\end{figure*}

To solve above mentioned problems, we proposed a comprehensive reasoning method for Doc-RE with knowledge retrieval augmentation and filtration, as shown in Figure \ref{fig2}. Firstly, a heterogeneous multi-level document graph was constructed for semantic encoding of entities, mentions, and sentences of the document. Then, a pre-trained co-reference resolution model was introduced to establish associations between entities and their corresponding pronouns for co-reference reasoning. Moreover, the document graph was extended with the retrieved external knowledge for common-sense reasoning. On this basis, a knowledge filtration method was proposed to determine whether to accept external knowledge. Finally, the axial attention method was integrated into KnowRA to realize logical reasoning across multi-sentences with intermediary entities.

The main contributions of our work are three folds:
\begin{itemize}
\item A comprehensive method for Doc-RE was proposed by achieving document semantic encoding with a multi-layer heterogeneous document graph, integrating the co-reference resolution model, injecting external knowledge, and introducing the axial attention mechanism.
\item A knowledge filtration method based on confidence score was presented for judging whether to accept external knowledge by filtering out irrelevant knowledge.
\item Extensive experiments performed on two public datasets demonstrated the superiority of our method compared to the state-of-the-art (SOTA) baselines.
\end{itemize}

\section{Our Methodology}


\subsection{Construction of document graph}

Firstly, pre-trained language models were employed to perform semantic encoding operations on the input document. 
Then, a multi-level heterogeneous document graph was present to model the connections among different entities, mentions, sentences in a document, defined as follows: 
\begin{definition}
 \label{def7}
    \textbf{Multi-level Heterogeneous Document Graph (MHDG)}. MHDG=$<V,E>$, where $V=\{v|v\in V^M \bigcup V^S \bigcup V^D\}$  represents the node set and $E 
    =\{<v_i,v_j>|v_i,v_j\in V ,i \neq j\}$, which represents edges between nodes $v_i$ and $v_j$. 
\end{definition}

According to definition \ref{def7}, we defined 3 types of node in MHDG: Mention node ($V^M$), Sentence node ($V^S$), and Document node ($V^D$). Then, four types of edges were given:
\begin{enumerate}[1)]
    \item \textbf{Document-Sentence Edge}: $E^{DS}=\{<v_i^D,v_j^S>| v_i^D\in V^D,v_j^S \in V^S\}$. The document node and all sentence nodes are connected through these edges.
    \item \textbf{Sentence-Sentence Edge}: $E^{SS}=\{<v_i^S,v_j^S>| v_i^S,v_j^S \in V^S, i\neq j\}$. Two adjacent sentences are connected by sentence-sentence edges. 
    \item \textbf{Mention-Sentence Edge}: $E^{MS}=\{<v_i^M,v_j^S>| v_i^M\in V^M,v_j^S \in V^S\}$, connecting mention nodes and the sentence node appearing with the same one sentence. 
    \item \textbf{Mention-Mention Edge}: Two mention nodes are connected by the Mention-Mention Edge (MME), which can be further divided into two categories: Co-Occurrence Mention-Mention Edge (CO-MME) means different mentions appear in the same sentence, formalized as: $E^{CO-MME}=\{<v_i^M,v_j^M>| v_i^M,v_j^M \in V^M\}$. Co-Reference Mention-Mention Edge (CR-MME) means that different mentions refer to the same entity, formalized as: $E^{CR-MME}=\{<v_i^M,v_j^M>| v_i^M,v_j^M \in V^M\}$.  
\end{enumerate}

Then, the semantic representation was performed on the MHDG with a graph-based neural network, denoted as:

\begin{equation}
\label{eq1}
\begin{aligned}
     H_{\{t_1,\dots,t_L\}}&=Encoder(\mathcal{T}_{input})\\
     &=Encoder({t_1,\dots,t_L})=[{h_1,\dots,h_L}]
\end{aligned}
\end{equation}
where $\mathcal{T}_{input}$ represents the input tokenized sequence, $L$ represents the length of the document. The representations of three kinds of nodes are as follows: 
\begin{align}\label{eq2}
    &H^D=h_{[CLS]}\\\label{eq3}
    &H^{m^i_j}=h_{P^{i,j}_e}=h_{``*"}\\\label{eq4}
    &H^{S_i}=\log\sum^{|S_i|}_{j=1}\exp{(h_j)}
\end{align}
where {$H^D\in R^d$} represents the $d$ dimension semantic representation of the document node. $H^{m^i_j}$ represents the semantic embedding of mention node for the \textit{j}-th mention of the \textit{i}-th entity, and $P^{i,j}_e$ represents the position of the \textit{j}-th mention of the \textit{i}-th entity, and “*” denotes for the special symbol placed before the mention $m^i_j$. $H^{S_i}$ represents the semantic representation of sentence node $S_i$.

Then, the Graph Attention Network (GAT) was used to calculate association score $AS(e_{i,j})$ for each edge between node \textit{i} and \textit{j}:
\begin{equation}\label{eq5}
    AS(e_{i,j})=LeakyReLU(W_\alpha[W_{\beta_1}h_i\oplus W_{\beta_2}h_j])
\end{equation}
where $\oplus$ represents the concatenation operation. The updated node representations based on MHDG are as follows:
\begin{align}\label{eq6}
    &h_i=\sum_{j \in N(i)}a_{i,j}(W_\beta h_j) \\
    \label{eq78}
    &a_{i,j}=softmax(AS(e_{i,j}))
\end{align}
where $W_\alpha \in R^d$ and $W_\beta \in R^{d\times d}$ are trainable parameters. $N(i)$ represents the number of adjacent nodes to node \textit{i}.

\subsection{Co-reference reasoning component for Doc-RE}

To solve the co-reference reasoning problem, we firstly pretrained a co-reference resolution model, $M_{coref}$, and then used it to identify the co-reference pronouns, as follows:
\begin{equation}\label{eq8}
    C_i=M_{coref}(\mathcal{T}_{input},E_i)=\{c^i_k\}^{n^c_i}_{k=1}
\end{equation}
where $C_i$ represents the set of identified co-reference pronouns and $n^c_i$ represents the number of the co-reference pronouns. Then, we used attention matrix between each input token and recognized co-reference pronouns to update semantic representations of co-reference pronouns, as follows:
\begin{align}
    \label{eq9}
    &A=MultiHeadAttention(Encoder(\mathcal{T}_{input}))\\
    \label{eq10}
    &Q_i=\sum^H_{h=1}(\frac{1}{n^c_i}\sum^{n^c_i}_{k=1}A^{h,k}_i)\\
    \label{equ12}
    &h_{c^i_k}=Q'_i[P^{i,k}_c]\cdot h_{P^{i,k}_c}
\end{align}
where $\textit{A}\in R^{H\times L\times L}$ is the attention matrix of all input tokens, $H$ represents the number of attention heads, and $L$ represents the length of input tokens. $Q_i\in R^L$, represents the averaged vector of attention value of entity $E_i$ to its co-reference pronouns. $A^{h,k}_i$ represents the \textit{h}-th attention head of entity $E_i$ to its \textit{k}-th co-reference word. $P^{i,k}_c$ represents the position of each co-reference pronoun.

Finally, we obtained the semantic representation of $c^i_k$ as shown in formula (11), where $Q'_i[P^{i,k}_c] \in R^{n^c_{i,k}}$ represents the normalized attention value of entity $E_i$ to its \textit{k}-th co-reference pronoun. $h_{P^{i,k}_c}$ represents the semantic embedding of the co-reference pronoun $c_{i,k}$. 

\subsection{Common-sense reasoning with the retrieval and filtration of document knowledge graph}

\subsubsection{Construction of knowledge-augmented document graph}
To introduce external knowledge for common-sense reasoning, we further constructed a knowledge-augmented document graph, denoted as K-MHDG.
\begin{definition}
    \textbf{Knowledge-retrieval-augmented Multi -level heterogeneous document graph (K-MHDG)}. K-MHDG$=<V,E\cup E_{know}>$, where $V=\{v|v \in V^{\{M,S,D\}}\}$ represents the entity node set, $E \subseteq \{<v_i,v_j>|v_{i,j}\in V ,i \neq j\}$ represents the relation edges between node $v_i$ and $v_j$ in MHDG, and $E_{know}$ represents the newly added edges, consisting of relations between entities, denoted as $r_{know}$, retrieved from the external knowledge base, formalized as: $E_{know} \subseteq \{<v_m,v_n>|\exists e^{r_{know}}=<v_m,v_n> \in hasRel(Q_{id}^{E_i},r_{know},Q_{id}^{E_j}), v_{m,n}\in V^M\}$ .
\end{definition}

We selected Wikidata as the external knowledge base to construct the K-MHDG. By utilizing the interface $getQid(EntityName)$ provided by Wikidata, all possible entity identifiers, i.e., $Qids$, were obtained based on the corresponding entity name. Then, based on the $hasRel(Q_{id}^{t},r_{know},Q_{id}^{h})$ interface, we can query all relation triples with the provided entity pairs. Finally, the newly retrieved relation $e^{r_{new}}=<v^M_{Ent_i},v^M_{Ent_j}>$ would be added to $E_{know}$ and the MHDG would be augmented to K-MHDG. 

\subsubsection{Semantic encoding for the external knowledge}

Based on K-MHDG, the representation of entity node $E_i$ was updated as follows:
\begin{align}\label{eq13}
h_{E_i}=\log \left( \frac{1}{n_{i}^{m}}\sum_{j=1}^{n_{i}^{m}}{\exp \left( h_{m_{j}^{i}} \right)}\oplus \frac{1}{n_{i}^{c}}\sum_{k=1}^{n_{i}^{c}}{\exp \left( h_{c_{k}^{i}} \right)} \right) 
\end{align}
where $\oplus$ represents the concatenation operation, $h_{m^i_j}$ represents the embedding of mention $m^i_j$ and $n^m_i$ represents the number of mentions related to $E_i$. $h_{c^i_k}$ and $n_i^c$ refer to the semantic embedding and number of the co-reference pronouns for entity $E_i$, respectively.


\subsubsection{External knowledge filtration method}
Considering that the external knowledge introduced is not always correct, we further proposed a new confidence-score-based knowledge filtration method to help our model autonomously determine whether to accept external knowledge.

The confidence score for each edge, i.e., $e_{i,j}=<E_i,r_k,E_j>$, was denoted as $\tau_{i,j}$ and calculated as:
\begin{equation}\label{eq16}
    \tau_{i,j}^k=f_{conf}(<E_i,r_k,E_j>)= h^T_{E_i}\cdot diag(r_k) \cdot h_{E_j}
\end{equation}
where $diag(\cdot)$ function represents the diagonal matrix with the vector $r_{k}$ as the diagonal element. The confidence score represents the likelihood of a relation existing between entity $E_i$ and $E_j$. After calculating the confidence score, the entity representation of entity $E_i$ was also updated as follows:
\begin{equation}\label{eq17}
    h'_{E_i}=\sum_{j \in N(i)}\sum_{k \in r_{i,j}} \sigma(\tau_{i,j}^k)(h_{E_j}\cdot r_k)
\end{equation}
where $\sigma(\cdot)$ represents the sigmoid function, which is used to convert confidence score into probabilities with values between 0 and 1. $r_{i,j}$ represents the set of relations that exist between entities $E_i$ and $E_j$.

Then, we used the correct relation label set provided by the annotated training set to train the model in a way that increases the confidence score for correct relations and decreases the confidence score for incorrect relations. The optimization function is:

\begin{equation}\label{eq18}
    \mathcal{L}_{KRA}=-\frac{1}{N}\sum^N_{n=1}(y_n\cdot \log(\sigma(\tau_{i,j}^k))+(1-y_n)\cdot \log(\sigma(1-\tau_{i,j}^k)))
\end{equation}
where $N$ represents the edge number in K-MHDG, and $y_n$ represents the annotated relation label in the training set, which value is 1 (correct relation) or 0 (incorrect relation).

\subsection{Logical reasoning with axial attention}

\subsubsection{Semantic fusion based on common context of entity pair}
Because Doc-RE is implemented in the entity pairs, it is necessary to integrate context information into entity pairs for logical reasoning.

We used the attention mechanism proposed to obtain the context of the common concern of entity $E_i$ and $E_j$, that is, the relevant context representation of entity pair $<E_i, E_j>$. The formula was given as follows: 
\begin{equation}\label{eq19}
    C^{i,j}=\frac{Q_i \times Q_j}{Q_i^T \cdot Q_j}H^D
\end{equation}
where symbol “$\times$” represents the outer product of embeddings, and “$\cdot$” represents the dot product of embeddings. 

Then, the semantic embedding for head and tail entities fused with the context information was calculated as:
\begin{align}\label{eq20}
    & z_i=\tanh(W_i h_{e_i}+W_c C^{i,j})\\
    \label{eq21}
    & z_j=\tanh(W_j h_{e_j}+W_c C^{i,j})\\
    \label{eq22}
    & z_{i,j}=z^T_iW_bz_j+b
\end{align}
where $z_i$ and $z_j$ represent the entity embedding fused with the local context information. $W_*\in R^d$ represents the trainable model parameters. Finally, the semantic representation of entity pair $<E_i,E_j>$, denoted as $z_{i,j}$, was obtained..

\subsubsection{ Axial-attention-based method for logical reasoning across sentences}

We arranged all entity pair representations $z_{i,j}$ in the document into an entity pair matrix of $N \times N$, where $N=|V|$ represents the number of entities in the document. If there is an intermediary entity $E_k$ between entity pair $<E_i,E_j>$, the purpose of our proposed method is to establish a multi-hop logical reasoning model by integrating semantic representations of all intermediary entities and the corresponding entity pairs.

In the entity pair matrix, the semantic information of the horizontal intermediary entity pair $<E_i,E_k| k=1,\dots,N>$ was firstly fused, calculated as follows:

\begin{equation}\label{eq23}
    G^{hor}_{<E_i, E_j>}=z_{i,j}+\sum_{k=1,\dots,N} softmax_k(q^T_{i,j}k_{i,k})v_{i,k}
\end{equation}
where $q_{i,j}, k_{i,j}, v_{i,j}=W_qz_{i,j}, W_kz_{i,j}, W_vz_{i,j}$ is the query, key, and value vector obtained by linear transformation of  $z_{i,j}$. $W_q, W_k, W_v \in R^{d \times d}$ are query, key, and value vectors obtained by linear transformation. 

Then, the similar operation was performed on all vertical entity pairs in the same column and calculated as follows:
\begin{equation}\label{eq24}
    G^{vert}_{<E_i, E_j>}=z_{i,j}+\sum_{k=1,\dots,N} softmax_k(q^T_{i,j}k_{k,j})v_{k,j}
\end{equation}

Finally, all intermediary entity pairs $<E_i,E_k>$ and $<E_k,E_j>$ were fused and the final representation for $<E_i,E_j>$ was obtained by :

\begin{equation}\label{eq25}
    G_{<E_i, E_j>}=G^{hor}_{<E_i, E_j>}+G^{vert}_{<E_i, E_j>}
\end{equation}

\subsubsection{Adaptive Relation Extraction Loss for Doc-RE}

Considering Doc-RE is a multi-label classification task, we adopted adaptive threshold loss function \cite{ref20,ref21} with a threshold class, denoted as TH, to adaptively separate positive ($\textbf{\textit{R}}^+$) and negative ($\textbf{\textit{R}}^-$) relations:
\begin{equation}\label{eq26}
    \begin{aligned}
        \mathcal{L} _{RE}=-\log \left( \frac{\exp \left( L_{i,j}^{TH} \right)}{\sum_{r^{\prime}\in \boldsymbol{R}^-\cup \left\{ TH \right\}}{\exp \left( L_{i,j}^{\prime} \right)}} \right) - \\ \sum_{i\ne j}{\sum_{r\in \boldsymbol{R}^+}{\log \left( \frac{\exp \left( L_{i,j}^{r} \right)}{\sum_{r^{\prime}\in \boldsymbol{R}^+\cup \left\{ TH \right\}}{\exp \left( L_{i,j}^{r\prime} \right)}} \right)}}
    \end{aligned}
\end{equation}
where $L_{i,j}$ represents the probability that entity pair $<E_i, E_j>$ belongs to each predefined relation type:
\begin{equation}\label{eq27}
    L_{i,j}=W_{l}G_{<E_i, E_j>}+b_l
\end{equation}

Through the joint training of the loss function $\mathcal{L}_{RE}$ and $\mathcal{L}_{KRA}$, the final optimization function for Doc-RE is:

\begin{equation}\label{eq28}
    \mathcal{L}=\mathcal{L}_{RE}+\lambda \cdot \mathcal{L}_{KRA}
\end{equation}
where $\lambda$ is the pre-defined hyper-parameter.

\section{Experiments and Analyses}

\subsection{Experimental Settings}

\textbf{Datasets}. We evaluated our model on two public datasets for document-level RE. \textbf{Re-DocRED} \cite{ref22} is a high-quality revised version of DocRED \cite{ref10}. Re-DocRED corrects the false negatives problem in dataset DocRED and contains 3,053 documents for training, 500 for development, and 500 for the test set. \textbf{DWIE} \cite{ref23} is sampled and annotated from the news website Deutsche Welle, containing 602, 98, 99 documents for training, development, and testing, respectively, with 43,373 entities, 21,749 relational facts, and 65 relation types.

\textbf{Evaluation metrics}. We used the micro F1 (\textbf{F1}), ignore F1 (\textbf{Ign F1}), Intra F1, and Inter F1 as the metrics for model performance, following previous work \cite{ref24}. \textbf{Ign F1} is a revised version of F1, which excludes the shared relations between the training and development/test set. \textbf{Intra F1} is used to evaluate F1 of relation triples that appear in the same sentence. \textbf{Inter F1} is used to evaluate F1 of cross-sentence relation triples.

\subsection{Baselines}

According to different model structures, the following models were selected for performance comparison.
\begin{itemize}
    \item Sequence-based models: These models used different deep neural structures for relation extraction, including convolution neural network (CNN), Bi-LSTM, and Context-Aware LSTM \cite{ref2}.
    \item Graph-based models: These models used graph neural network for Doc-RE, including GAIN \cite{ref26}, SIRE \cite{ref28}, DRN \cite{ref27}, SagDRE \cite{ref12}.
    \item Transformer-based models: These Transformer-based models \cite{ref29} include SSAN \cite{ref30}, ATLOP \cite{ref20}, ATLOP-MILR \cite{ref50}, DocuNET \cite{ref31}, KD-DocRE \cite{ref32}, UGDRE \cite{ref52}, and JMRL-DREEAM \cite{ref49}. It should be noted that documents in dataset DWIE are quite long, with over 50\% of documents exceeding 768 tokens, and the maximum length reaches 2560 tokens. Therefore, for dataset DWIE, we replaced RoBERTa\textsubscript{\textit{-large}} with LongFormer \cite{ref34} as the backbone model, which supports a maximum of 4096 tokens. For dataset Re-DocRED, baseline models used RoBERTa\textsubscript{\textit{-large}} as the cornerstone model.
    \item LLM-based models: We used 13B LLAMA-2\footnote{https://github.com/meta-llama/llama.} model as the large language model for Doc-RE, finetuned with LoRA\footnote{https://github.com/microsoft/LoRA.}.
\end{itemize}

Considering that these baseline models used different datasets and cornerstone models in their original papers, we tried our best to rerun and fine-tune these models for fair comparison. For example, the rerun model JMRL-DREEAM even achieved better performance on dataset Re-DocRED compared to its original paper.
\subsection{Main Results}

Table 1 listed the performance of models on two datasets. We observed that: 1) Our KnowRA model outperformed other baseline models in almost metrics on two datasets. For dataset Re-DocRED and DWIE, KnowRA surpassed the existing SOTA model, JMRL-DREEAM, by 0.28 and 1.13 in F1 score, respectively. These experiments proves the superiority of our model for Doc-RE. 2) In terms of Intra- and Inter-F1, our model achieved the second-best results in dataset Re-DocRED and the best performance in dataset DWIE. These experimental results proved the advantages of our model in cross-sentence relation extraction. Meanwhile, for dataset DWIE with longer document length, our method achieved better performance compared to the SOTA models, i.e., KD-DocRE and JMRL-DREEAM, which indicates that our model has stronger semantic reasoning ability for extracting long-distance document-level relations. 3) For both datasets, our model outperformed the LLaMA-2-based model in all metrics,  which implies that large language models need to improve their effectiveness through relevant optimization technologies when applied to downstream tasks.

\begin{table*}[!t]
\label{table1}
\centering
\small

\begin{tabular}{l|llll|llll}
\hline
\multicolumn{1}{c|}{}       & \multicolumn{4}{c|}{\textbf{Re-DocRED}}                                                                                                                                               & \multicolumn{4}{c}{\textbf{DWIE}}                                                                                          \\ \cline{2-9} 
\multicolumn{1}{c|}{\multirow{-2}{*}{\normalsize \textbf{Models}}} & \multicolumn{1}{c}{Ign F1}               & \multicolumn{1}{c}{F1}                   & \multicolumn{1}{c}{Intra-F1}                & \multicolumn{1}{c|}{Inter-F1}            & \multicolumn{1}{c}{Ing F1} & \multicolumn{1}{c}{F1} & \multicolumn{1}{c}{Intra-F1} & \multicolumn{1}{c}{Inter-F1} \\ \hline
CNN \cite{ref10}                              & 54.28\textsubscript{±0.33}                            & 56.20\textsubscript{±0.35}                               & 59.61\textsubscript{±0.62}                                  & 53.54\textsubscript{±0.67}                               & 40.24\textsubscript{±0.19}                 & 51.84\textsubscript{±0.19}             & 51.84\textsubscript{±0.19}                   & 51.31\textsubscript{±0.63}                   \\
BiLSTM \cite{ref10}                                       & 58.08\textsubscript{±0.38}                               & 60.03\textsubscript{±0.30}                               & 62.99\textsubscript{±0.07}                                  & 57.70\textsubscript{±0.51}                               & 55.07\textsubscript{±0.18}                 & 66.21\textsubscript{±0.25}             & 68.58\textsubscript{±0.28}                   & 64.78\textsubscript{±0.38}                   \\
Context-Aware \cite{ref10}                                 & 58.29\textsubscript{±0.26}                               & 60.19\textsubscript{±0.21}                               & 63.18\textsubscript{±0.32}                                  & 57.82\textsubscript{±0.17}                               & 56.80\textsubscript{±0.18}                 & 66.05\textsubscript{±0.17}             & 69.24\textsubscript{±0.39}                   & 63.54\textsubscript{±0.43}                   \\ \hline
GAIN \cite{ref26}                                          & 73.55\textsubscript{±0.15}                               & 74.89\textsubscript{±0.21}                               & 77.46\textsubscript{±0.31}                                  & 72.68\textsubscript{±0.15}                               & 63.49\textsubscript{±0.57}                 & 68.62\textsubscript{±0.32}             & 68.97\textsubscript{±0.34}                   & 68.28\textsubscript{±0.43}                   \\
SIRE \cite{ref28}                                        & 73.10\textsubscript{±0.40}                               & 74.55\textsubscript{±0.38}                               & 77.28\textsubscript{±0.46}                                  & 72.22\textsubscript{±0.63}                               & 63.01\textsubscript{±0.27}                 & 68.31\textsubscript{±0.22}             & 68.07\textsubscript{±0.29}                   & 67.74\textsubscript{±0.37}                   \\
DRN \cite{ref27}                                           & 72.37\textsubscript{±0.23}                               & 73.28\textsubscript{±0.22}                               & 76.28\textsubscript{±0.15}                                  & 70.58\textsubscript{±0.34}                               & 63.13\textsubscript{±0.32}                 & 69.32\textsubscript{±0.23}             & 71.51\textsubscript{±0.23}                   & 67.07\textsubscript{±0.38}                   \\
SagDRE \cite{ref12}                                       & 73.44\textsubscript{±0.29}                               & 74.56\textsubscript{±0.23}                               & 76.99\textsubscript{±0.18}                                  & 72.46\textsubscript{±0.38}                               & 63.37\textsubscript{±0.27}                 & 69.61\textsubscript{±0.31}             & 69.84\textsubscript{±0.26}                   & 68.98\textsubscript{±0.35}                   \\ 
 \hline

SSAN \cite{ref30}                                         & 72.64\textsubscript{±0.32}                               & 73.88\textsubscript{±0.28}                               & 75.28\textsubscript{±0.38}                                  & 72.20\textsubscript{±0.35}                               & 76.26\textsubscript{±0.24}                & 81.06\textsubscript{±0.10}             & 86.10\textsubscript{±0.24}                   & 77.09\textsubscript{±0.39}                   \\
ATLOP \cite{ref20}                                        & 76.85\textsubscript{±0.29}                               & 77.48\textsubscript{±0.30}                               & 79.54\textsubscript{±0.28}                                  & 75.65\textsubscript{±0.34}                               & 78.67\textsubscript{±0.24}                 & 83.21\textsubscript{±0.19}             & 87.25\textsubscript{±0.11}                   & 80.84\textsubscript{±0.32}                   \\
ATLOP-MILR \cite{ref52}                                        & 75.99\textsubscript{±0.24}                               & 76.68\textsubscript{±0.17}                               & 78.95\textsubscript{±0.21}                                  & 74.69\textsubscript{±0.19}                              & 79.95\textsubscript{±0.29}                 & 84.66\textsubscript{±0.41}             & 87.04\textsubscript{±0.68}                   & 82.29\textsubscript{±0.14}                  \\
DocuNET \cite{ref31}                                       & 77.19\textsubscript{±0.22}                               & 77.88\textsubscript{±0.26}                               & 79.89\textsubscript{±0.21}                                  & 76.11\textsubscript{±0.41}                               & 79.41\textsubscript{±0.21}                 & 84.18\textsubscript{±0.13}             & 87.88\textsubscript{±0.18}                   & 80.84\textsubscript{±0.32}                  \\
KD-DocRE \cite{ref32}                                      & 77.34\textsubscript{±0.33}                               & 78.12\textsubscript{±0.30}                               & 80.19\textsubscript{±0.29}                                  & 76.31\textsubscript{±0.35}                               & 80.22\textsubscript{±0.24}                 & 84.87\textsubscript{±0.19}             & \underline{88.01\textsubscript{±0.35}}                   & 81.57\textsubscript{±0.32}                   \\

UGDRE \cite{ref52}                     & 78.15\textsubscript{±0.30}  & 78.87\textsubscript{±0.27}               & \textbf{81.05\textsubscript{±0.24}}          & 76.93\textsubscript{±0.35} &     72.08\textsubscript{±0.59}       &  76.35\textsubscript{±0.61}          &  79.51\textsubscript{±0.72}        & 73.44\textsubscript{±0.60}             \\

JMRL-DREEAM   \cite{ref49}                                          & \textbf{78.57\textsubscript{±0.06}}                              & \underline{78.97\textsubscript{±0.06}}                               & 80.13\textsubscript{±0.29}                                  & \textbf{77.96\textsubscript{±0.18}}                               & \underline{80.60\textsubscript{±0.55}}                & \underline{85.27\textsubscript{±0.36}}            & 87.88\textsubscript{±0.34}                   & 82.68\textsubscript{±0.72}                  \\ \hline
LLaMA-2                                       & 46.84\textsubscript{±0.40}                               & 47.02\textsubscript{±0.43}                               & 45.66\textsubscript{±0.88}                                  & 72.42\textsubscript{±0.29}                               & 80.57\textsubscript{±0.69}               & 82.37\textsubscript{±0.58}             & 79.74\textsubscript{±0.88}                   & \underline{82.84\textsubscript{±0.31}}                   \\ \hline
KnowRA \textbf{(ours)}                                 & \underline{78.39\textsubscript{±0.32}}                      & \textbf{79.25\textsubscript{±0.16}}                      & \underline{ 80.42\textsubscript{±0.33}}                            & \underline{76.99\textsubscript{±0.28}}                      & \textbf{81.48\textsubscript{±0.32}}        & \textbf{86.40\textsubscript{±0.22}}    & \textbf{88.17\textsubscript{±0.09}}          & \textbf{83.83\textsubscript{±0.45}}          \\ \hline
\end{tabular}
\caption{Model performance in two datasets. \textbf{Bold} denotes the best result, \underline{underline} denotes the second best result. We ran each experiment five times, using different random seeds, and reported their performance and standard deviation.}
\end{table*}

\subsection{Ablation Experiments}

Ablation experiments were shown in Table 2. For dataset Re-DocRED, after removing the document graph (denoted as “\textit{-w/o graph}”) and axial attention (denoted as “\textit{-w/o axial atten.}”), the model performance decreased by 1.07 and 1.01 in F1 score, respectively. When both the two components were removed (denoted as “\textit{-w/o graph+axial}”), the performances on F1 decreased by 1.28. The removal of the co-reference reasoning component (denoted as “\textit{-w/o coref}”) also leads to a drop of 0.70 in terms of F1. 

In addition, the common-sense reasoning component based on the knowledge-retrieval-augmentation (denoted as “\textit{-w/o knowAug.”}, representing the removal of the K-MHDG) is an most important component, as the F1 score decreased the most (-1.10 on Re-DocRED and -1.02 on DWIE) when it was removed. Also, when the knowledge filtration method (denoted as “\textit{-w/o knowFilt.}”) was removed from our complete model, the F1 score decreased by 0.97 and 0.79 on dataset Re-DocRED and DWIE, respectively, which proves that the filtration of the external knowledge is necessary for filtering out noise information and boosting performance. For dataset DWIE, we observed similar phenomena, which verifies the effectiveness of our comprehensive reasoning components.

\begin{table}[!t]
  \centering
\small  
  \label{table3}
\begin{tabular}{l|cc|cc}
\hline
\multicolumn{1}{c|}{\multirow{2}{*}{\textbf{Variant Models}}} & \multicolumn{2}{c|}{\textbf{Re-DocRED}} & \multicolumn{2}{c}{\textbf{DWIE}} \\ \cline{2-5} 
\multicolumn{1}{c|}{}                                         & \multicolumn{2}{c|}{\textbf{F1}}        & \multicolumn{2}{c}{\textbf{F1}}   \\ \hline
\textbf{KnowRA}                                               & \textbf{79.25}        & $\Delta$        & \textbf{86.40}     & $\Delta$     \\
\textit{  -w/o graph}                                           & 78.18                 & -1.07           & 85.65              & -0.75        \\
\textit{  -w/o coref}                                           & 78.55                 & -0.70           & 86.02              & -0.38        \\
\textit{  -w/o knowAug.}                                        & 78.15                 & -1.10           & 85.38              & -1.02        \\
\textit{  -w/o knowFilt.}                                       & 78.28                 & -0.97           & 85.61              & -0.79        \\
\textit{  -w/o axial atten.}                                    & 78.24                 & -1.01           & 85.57              & -0.83        \\
\textit{  -w/o graph+axial}                                     & 77.97                 & -1.28           & 85.25              & -1.15        \\ \hline
\end{tabular}
\caption{Ablation experimental results in Re-DocRED and DWIE.}
\end{table}

\subsection{Effects of Intra- and Inter-sentence Reasoning}

\begin{figure}[!htb]
     \centering
     \includegraphics[width=0.9\linewidth]{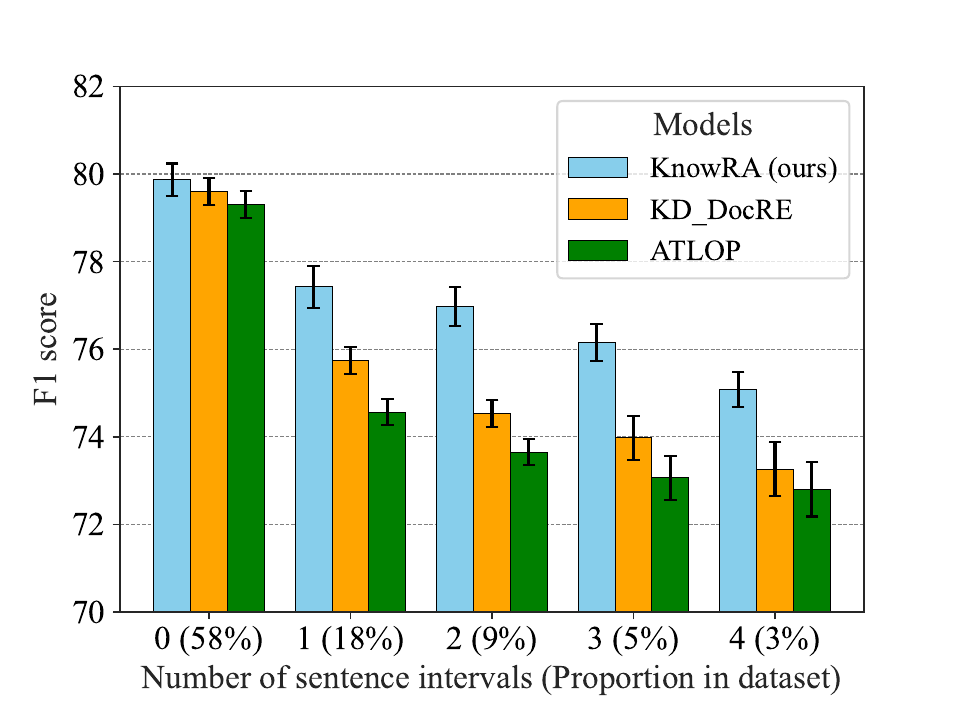}
     \caption{Model performance for Doc-RE changing with the length of sentence intervals on dataset Re-DocRED. For example, 1 (18\%) represents the case where the length of interval sentences between entities in a relation is 1, and its proportion in the dataset is 18\%.}
     \label{fig3}
 \end{figure}

The number of sentence intervals was used to present the distance between the head and tail entity in a relation triple. According to the results shown in Figure \ref{fig3}, we found that: 1) When the head and tail entities are in the same sentence (intervals = 0), our model performed better than the baselines. 2) When the head and tail entities are from different sentences (intervals $>$ 0), our model was also superior to other baselines. 3) The performance of all models gradually decreased with the increase of sentence intervals. This phenomenon indicates that our model performed well in both intra- and inter-senetence reasoning. 

 \subsection{Effects of Context Length}

 We conducted experiments to quantitatively evaluate the impact of the context length on the performance of Doc-RE. For dataset DWIE, the length of most document (94.81\%) is much greater than 512, which exceeds the maximum length supported by the RoBERTa\textsubscript{large}-based encoder. To this end, we replaced the original encoders of all compared models with the longFormer \cite{ref34}.  

\begin{figure}[!htb]
     \centering
     \includegraphics[width=0.8\linewidth]{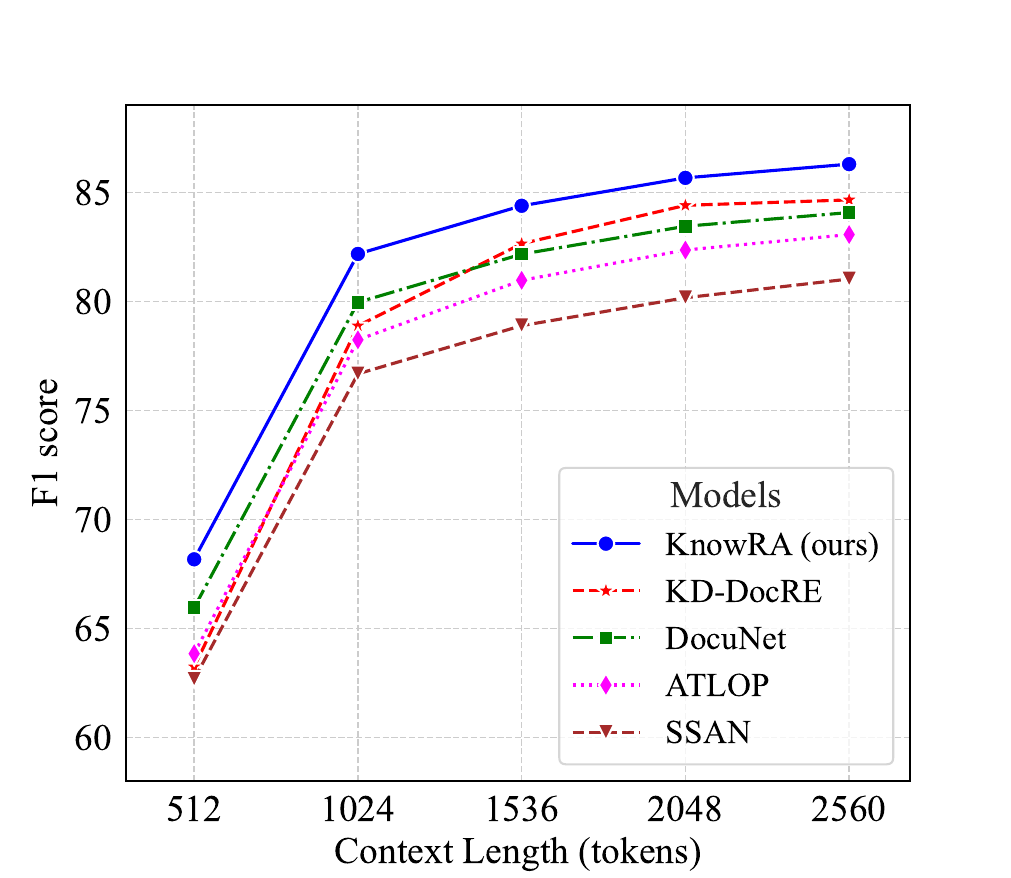}
     \caption{The trend of model performance changing with context length in the DWIE dataset}
     \label{fig4}
 \end{figure}

The experimental results are shown in Figure \ref{fig4} and we found that: 1) As the context length gradually increases, the performance of all models first improves, and then the growth rate gradually slows down. Similar trends were also observed on other baselines. These results proved that our model can outperform the SOTA models in different context length, which indicates a good scalability for our model.

\subsection{Impact of Knowledge-augmented Method}
We conducted a hyper-parameter analysis on weight $\lambda$ of the $\mathcal{L}_{KRA}$ loss function to quantitatively evaluate the impact of the proposed knowledge-retrieval-augmentation method. 

\begin{figure}
    \centering
    \includegraphics[width=\linewidth]
    {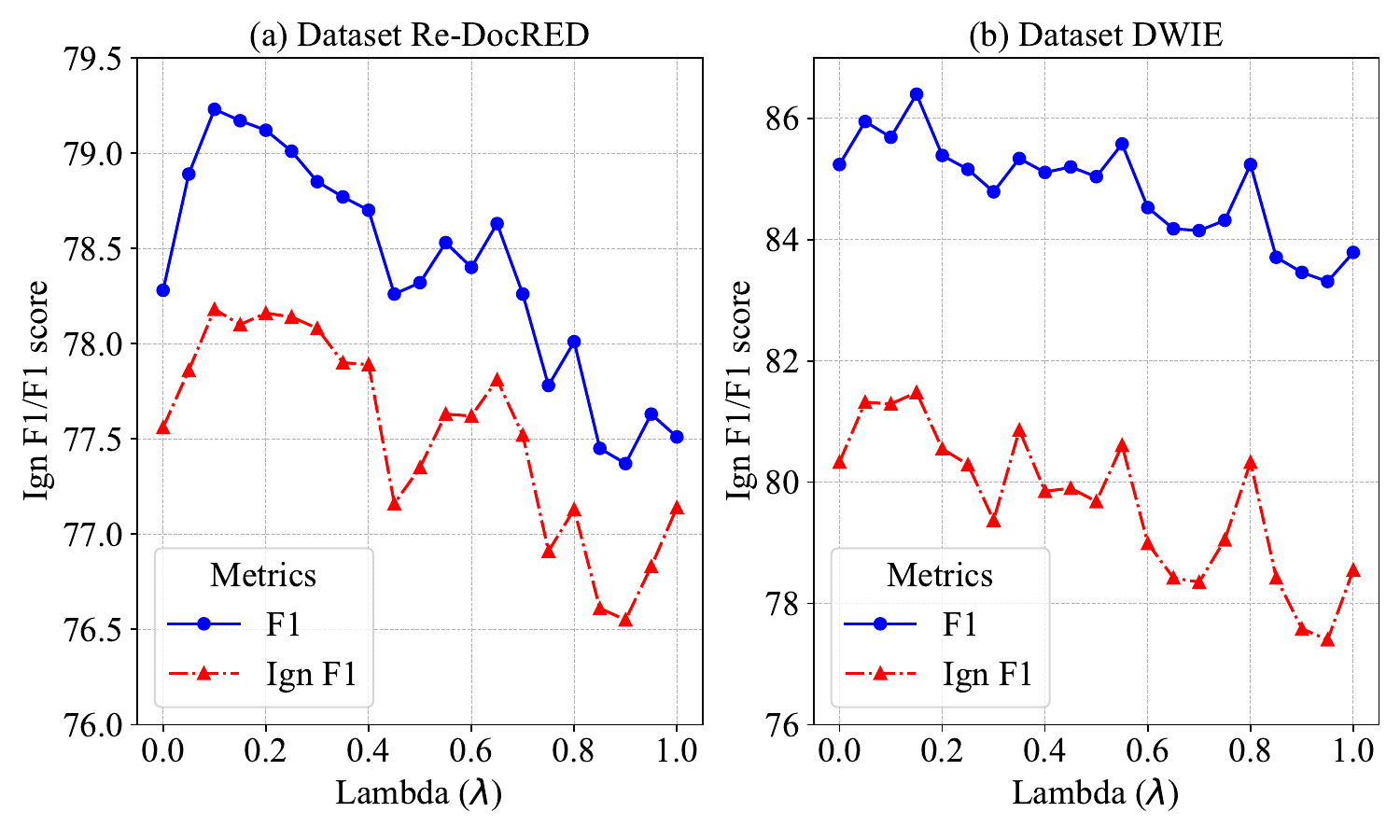}
    \caption{The trend of our KnowRA model performance changing with the weight $\lambda$ in datasets Re-DocRED and DWIE.}
    \label{fig5}
\end{figure}

As shown in Figure \ref{fig5}, it can be observed that: 1) For both datasets, the model performance shows a trend of first improving and then decreasing with the increase of parameter $\lambda$. 2) Introducing external knowledge through our proposed knowledge augmentation method can improve the model performance for Doc-RE. However, too high the weight value $\lambda$ would lead to a decline for the model performance. One possible explanation is that external knowledge may be wrong or inconsistent with the internal information of the document, which impairs the model performance.

\subsection{Case Study}
\begin{figure}[!htb]
    \centering
    \includegraphics[width=0.8\linewidth]{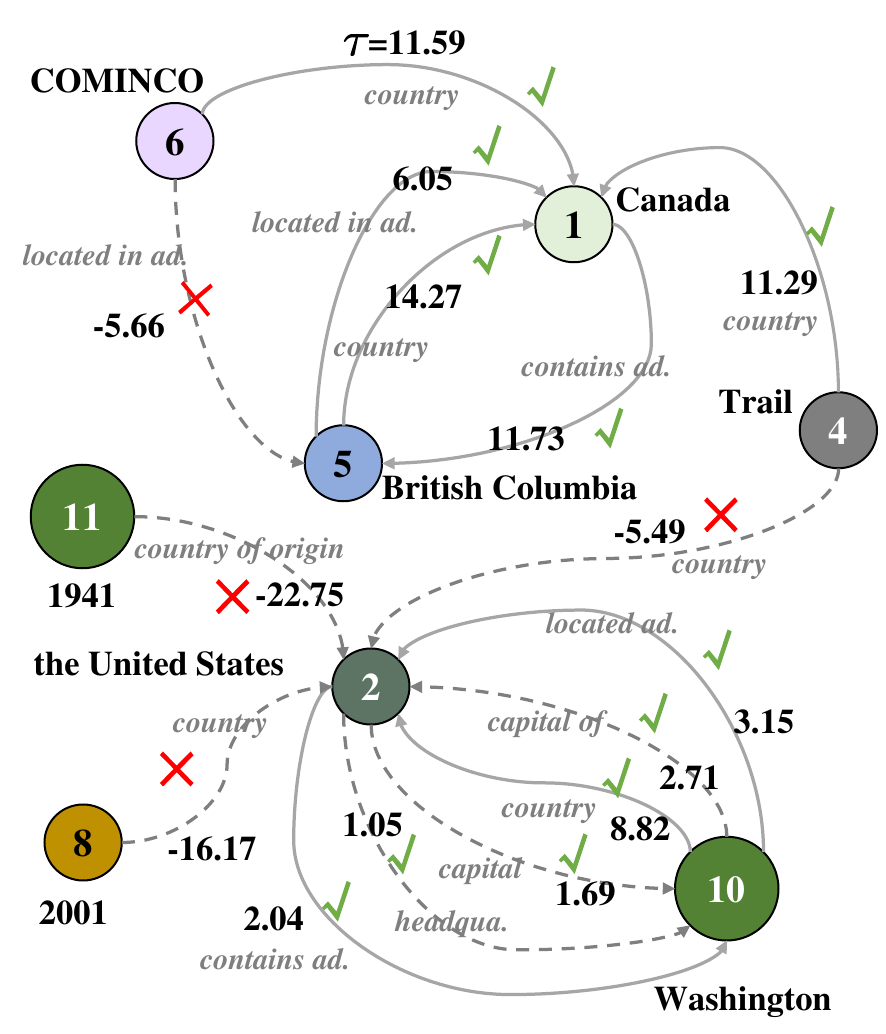}
    \caption{The illustration of K-MHDG, which is extended by external knowledge. The solid lines denote relation connections constructed based on labels from the original dataset, while dashed lines denote relation connections extended from external knowledge.}
    \label{fig6}
\end{figure}

Figure \ref{fig6} shows the construction of K-MHDG by introducing the common-sense knowledge from the external knowledge base. The nodes in the figure represent entities, where the numbers indicate the identification index of the entities. The directed edges between nodes represent the relations between entities, the words on the edges represent the relation types, and the numbers on the edges represent the confidence score, i.e., $\tau$, of the relation triples. 

Our knowledge augmentation method can extend the relation triples that were not annotated in the original dataset, identify the correct relation triples, denoted by the green \textcolor{green}{$\checkmark$}, and also filter out noise external knowledge and incorrect relation triples, which are denoted by the red \textcolor{red}{$\times$}. 

\section{Related Work}
For pattern recognition and logical reasoning, existing models mainly used entities/mentions as intermediary nodes to construct document graphs to achieve logical reasoning of cross-sentence relation triples \cite{ref9,ref36,ref12}.  Peng et al. \shortcite{ref13} proposed a subgraph seasoning model for Doc-RE, which places emphasis on key entities and integrates various paths between entity/mention pairs into a subgraph for relation reasoning. 

For co-reference reasoning, this type of model has positive effects on improving the performance of modeling long-distance reasoning by reducing ambiguity between co-reference entities and mentions involving multiple sentences \cite{ref40}. Ye et al. \shortcite{ref18} proposed mention reference prediction method to equip the language model with the capacity for capturing and representing the co-reference relations. Wang et al. \cite{ref19} proposed a co-reference distillation method, which distills the co-reference reasoning ability into the relation extraction model.

However, for common-sense reasoning, using only the information in the current document makes it difficult to establish implicit associations and determine relation types between different entities/mentions in multiple sentences, without the help of common sense distilled from external knowledge \cite{ref48}. Existing models rarely incorporate external knowledge into the RE model, enabling the model to possess relevant background knowledge like humans and reducing the difficulty of relation extraction across multi-sentences, which is still a tricky challenge for Doc-RE. To solve this problem, we introduced external knowledge into our model with the knowledge filtration method. 

\section{Conclusions}

In this paper, we presented a comprehensive reasoning reasoning model for Doc-RE. Concretely, the proposed knowledge document graph and three different reasoning components, i.e., graph-based semantic encoding, co-reference resolution, knowledge augmentation and filtration, and axial attention method were integrated into our KnowRA model to enhance the comprehensive reasoning abilities. Experiments conducted on two benchmarks datasets demonstrated the superiority of the proposed model in Doc-RE.

\appendix



\section*{Acknowledgments}
This work was supported by the National Natural Science
Foundation of China (Nos. 61572250 and 62476135), Jiangsu
Province Science \& Tech Research Program (BE2021729),
Open project of State Key Laboratory for Novel Software
Technology, Nanjing University (KFKT2024B53), Jiangsu
Province Frontier Technology Research and Development
Program (BF2024005), Nanjing Science and Technology Research Project (202304016) and Collaborative Innovation Center
of Novel Software Technology and Industrialization, Jiangsu,
China.


\bibliographystyle{named}
\bibliography{ijcai25}

\end{document}